# Mask-GD Segmentation Based Robotic Grasp Detection *

Mingshuai Dong[1], Shimin Wei[2], Xiuli Yu[3], Jianqin Yin[*]

*Abstract*— The reliability of grasp detection for target objects in complex scenes is a challenging task and a critical problem that needs to be solved urgently in practical application. At present, the grasp detection location comes from searching the feature space of the whole image. However, the cluttered background information in the image impairs the accuracy of grasping detection. In this paper, a robotic grasp detection algorithm named MASK-GD is proposed, which provides a feasible solution to this problem. MASK is a segmented image that only contains the pixels of the target object. MASK-GD for grasp detection only uses MASK features rather than the features of the entire image in the scene. It has two stages: the first stage is to provide the MASK of the target object as the input image, and the second stage is a grasp detector based on the MASK feature. Experimental results demonstrate that MASK-GD's performance is comparable with state-of-the-art grasp detection algorithms on Cornell Datasets and Jacquard Dataset. In the meantime, MASK-GD performs much better in complex scenes.

## I. INTRODUCTION

Grasping detection is an important research direction in the field of robotics. In pursuit of expanding the application scenarios of robots in our daily lives, it is essential for a robot to grasp an object in complex scenarios. The human being can achieve accurate and stable grasping in a complex scene because they focus on the feature of the object itself. Therefore, to grasp an object, robots need to find the feature of the object first. However, the cluttered background information makes the grasp detector challenging to extract the target feature. Thus, it is necessary to find a more accurate way to extract the feature of the object in grasp detection. In this paper, we propose a new approach to detect grasp locations for parallel plate gripper in a more accurate way.

Recent works [1][2][3][4] focus on grasp detection on the feature of the whole image. The proportion of background information in the image, however, is much higher than the target object. Therefore, the grasp detector will consume many computing resources in detecting background information, and the complex background information impairs the accuracy of grasp detection, as shown in Fig 1 (a). On the other hand, we cannot determine whether the model focuses on object features or background features in the grasp detection process. [5]try to solve the problem of grasping an object in complex scenes, which reduces the background information by extracting the ROI of the image. However, due to the ROI region's axisymmetric nature, the ROI region of irregular target object also contains background information or local information of other objects. Therefore, [5] does not eliminate the interference of background information on grasping accuracy.

In this paper, a new approach of robotic grasp detection based on MASK named MASK-GD is proposed. MASK is a segmented image, and the pixels of background information

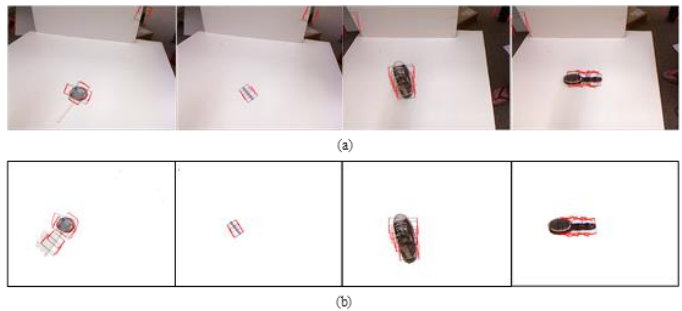

Figure 1 (a): the cluttered background will lead the grasp detector to generate a false grasp representation; (b): grasp detection results after deleting the background scene through target segmentation.

are deleted. MASK-GD includes two stages: the first stage is to provide MASK from the input image, and the second stage is a grasp detector based on MASK. Therefore, MASK-GD only uses the target object's features for grasping detection, which avoids the inference of background information to grasping detection.

In summary, the contributions of our work are as follows:

We propose a new grasp detection algorithm based on image segmentation, and the algorithm optimizes scene-oriented detection to object-oriented detection. Grasping detection is locally based on the feature of detailed foreground information, which is segmented from the whole image. Therefore, our method significantly reduces the range of grasp detection. We verify our method on Cornell Datasets and Jacquard Dataset, and it is comparable with state-of-the-art grasp detection algorithms.

We propose a new grasp detection network MASK-GD, which can simultaneously detect targets and grasps in multi-object scenes. Therefore, the network can grasp specified objects in multi-object scenarios.

The influence of the correlation between foreground information and background information on grasping detection is studies. And a new network solving grasping detection named Mask-GD is provided. It can realize objects segment of grasping object and background information while finding the target from multiple objects. At present, most of the research on grasping detection focuses on the modeling of the relationship among image pixels and ignores the contribution of foreground and background information to the grasping box. The results of this study show that our proposed model can eliminate the interference of background elements to grasp detection, and the performance of our model outperforms the current state-of-the-art algorithm.

## II. RELATED WORK

Robotic grasping detection has been studied for a long time, and its accuracy and efficiency have been improved significantly. At first, grasp detection widely adapted 3-D modeling method to find the grasp position of objects [11][12]. This method could achieve good grasp detection. This effective method performs satisfactorily, but it relies on the complete 3-D model and other object's physical information. However, the full 3-D models of objects are difficult to obtain, and general-purpose robots do not have the ability to build complex and comprehensive 3-D models of objects.

Recently, the convolutional neural network has a strong ability of feature extraction. [13]uses a convolutional neural network as a classifier in the sliding window detection pipeline to realize object grasping detection and finally achieved a satisfactory performance. However, the sliding window detection pipeline typically required moving the classifier numerous times on small patches of an image, which consumes a high computational cost. [1][1]proposed a convolutional neural network with a full connection layer to directly perform classification and regression operations on the whole RBG or RGB-D image to realize grasp detection, which exceeded the method proposed by [13]in recognition accuracy and real-time performance and dramatically reduce the calculation cost. At that time, the model finally achieved state-of-the-art performance on the Cornell Grasp Datasets. [4]successfully applied the model trained in the Cornell Grasp Dataset to multi-objective scenarios and achieved a good performance. However, the above method is frequently used to detect the whole image, and the grasping detection network needs to detect every part of the input image, which still lacks pertinence and low recognition efficiency. This method goes against the essentially human habit of object-oriented grasping.

[5]proposes an ROI-based grasp detection framework ROI-GD, which significantly reduced the area of grasp detection and improves the efficiency of grasp detection. The framework extracts the Region of Interest (ROI) in the input image through the RPN network and performs grasp detection on the ROI features. ROI-GD reduces the range of grasp detection and achieves excellent performance in multi-target and single-target scenarios. However, ROI is an axisymmetric region, and for irregular objects, there is a large amount of background information or local information about other targets in the ROI region. Therefore, ROI-GD doesn't implement object-oriented grasp detection, and it just shrinks the area of grasp detection.

Under the above conditions, an accurate grasping detection area is needed to realize object-oriented fetching detection. This paper proposes a robotic grasp detection method based on image segmentation, which can accurately extract the grasp detection area through image segmentation. The method we present achieved competitive results in both single-objective and multi-objective scenarios.

## III. PROBLEM DESCRIPTION

Currently, there are two alternative methods for the representation of grasp detection. One is the 7-dimensional grasp representation proposed by [14], which is mainly aimed at grasping 3-D objects. The other is a 5-dimensional grasping representation method for 2-D objects submitted by [13]. This method simplifies the 7-dimensional grasping representation [14] and can fully describe the relative position, direction, and opening width of the parallel plate gripper. Therefore, the 5-dimensional grasp representation method proposed by [13] is widely used in [1][3][4][15] and other grasp detection methods. In our work, we also use the 5 - dimensional representation. The 5-dimensional grasp is represented as follow：

$$g=\{x, y, \theta, h, w\} \quad (1)$$

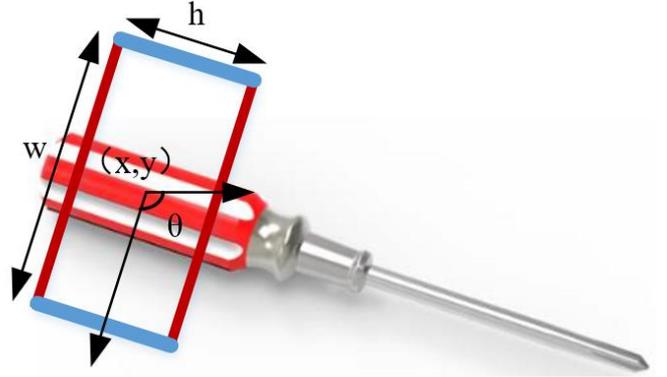

Figure 2 A 5D detection representation with location $(x, y)$, rotation $\theta$, gripper opening with $w$ and plate size $h$

Where $(x, y)$ is the center of the parallel plate gripper, $h$ is the height, and $w$ is the width. $\theta$ is the angle between the movement direction and the horizontal axis of the parallel plate gripper. Figure 2 shows an example of this grasp representation.

## IV. METHOD

### A. Network Architecture

At present, the grasping detection methods proposed by [4][15] and others are all whole-image-oriented, which are subject to interference from background information and require a large amount of calculation. These methods differ significantly from human grasping habits. [3]proposed to narrow the scope of grasp detection by extracting ROI. However, due to the irregularities of objects and the axisymmetric nature of the ROI region, background information, and local information of other objects would still exist in the ROI region. Therefore, this method does not realize object-oriented grasp detection. In this paper, we propose to use the instance segmentation method to distinguish the target and the background. Simultaneously, this method replaces the background information that significantly affected the grasp detection with pure white. This method eliminates the interference of background information and realizes the object-oriented grasp detection. Figure 3 shows the complete structure of the MASK-GD network.

To get the data of a specific target object for grasping detection, we detect the target in the scene simultaneously as target segmentation. The data are containing only the target object for grasp detection.

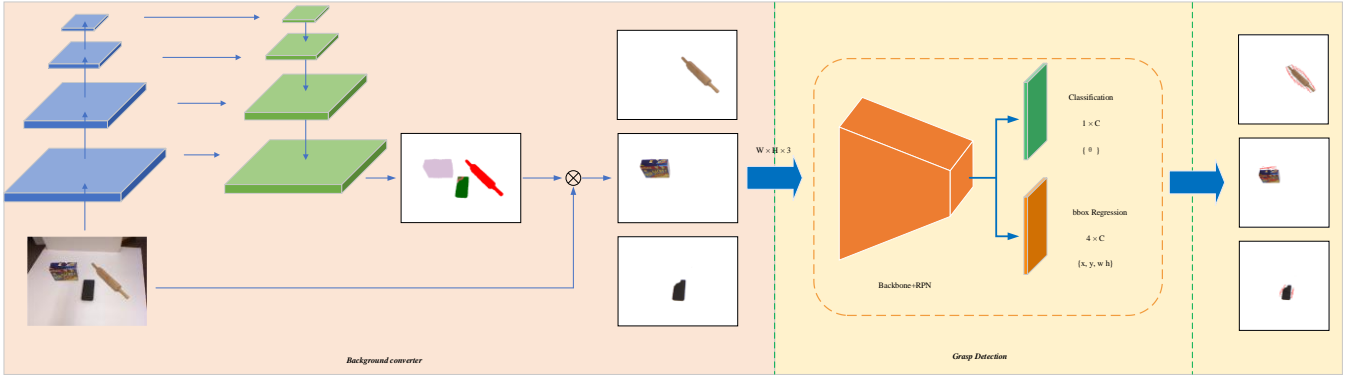

Figure 3. The complete structure of our grasp detection model is based on target segmentation. The model's input is an RGB image, and the model is divided into two stages: target segmentation stage and grasp detection stage. In the first stage, only the pixels of the target area is retained. In the second stage, the image containing only the target is grasp detected.

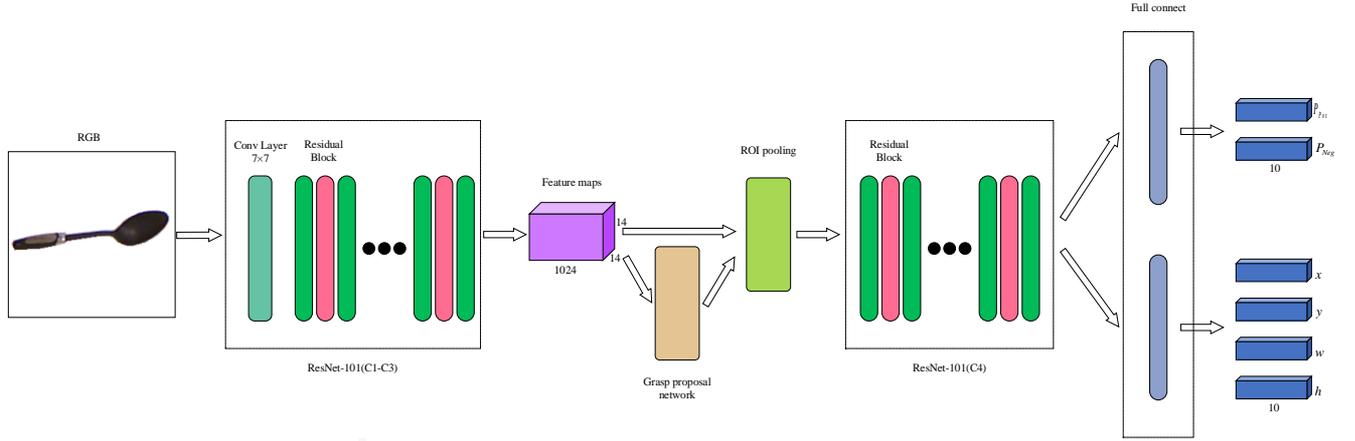

Figure 4. The structure of our multi-grasp predictor. The backbone network is ResNet-101.

Our method adopts a two-stage network structure. The first stage is the instance segmentation stage, in which the input image is segmented to distinguish the target object from the background information reliably. The second stage is the grasp detection stage, which only contains the target image to grasp detection.

In the first stage, inspired by the U-Net++[17] network, we use an encoder, decoder, and skip connection structure to segment the input image. The encoder adopts the VGG16[16] network as the backbone to extract the input image features and conduct four times down-sampling operations. The down-sampling process can significantly increase the robustness of the model to some perturbations of the input image, increase the sensitivity field of the model, and reduce the cost of computation and the risk of over-fitting. The decoder restores the feature map to the initial size through four up-sampling operations and concatenates the feature map generated by each down-sampling through a skip connection. The information loss caused by down-sampling is reduced, and the segmentation performance of the model for objects of different sizes is significantly improved using up-sampling and skipping layer connection.

After the image segmentation is completed, the model replaces the pixels in the target area with the pixel values corresponding to the original image based on the Mask segmented and fills the rest of the background with pure white, shown in Figure 4 and Figure 5. By this method, the target's texture information is retained to the greatest extent, and the interference of messy background information is effectively removed so that the real texture information of the object can be learned in the grasp detection stage.

Finally, after segmentation, the RGB image is input into the second-stage grasp detector, and the grasp detector performs grasp detection against the target in the image. The experimental verification shows that our method can effectively improve the grasping accuracy, and the grasping result is better than the ROI-GD method proposed by [3]. The method we present can be successfully applied to multi-objective scenarios, while state-of-the-art results can still be obtained in single-objective scenarios.

The next three sections describe the overall architecture of the system. It includes object-oriented grasping detection, the relationship with the proposed grasping detection algorithm, and the loss function.

### B. Object - Oriented Multi-Grasp Detection

In this paper, to make the grasp detection only target object rather than the whole image, we segment the target object and background information by image segmentation method and replace the pixels in the area unrelated to the target object so that only the pixel information of the target object exists in the detection area.

In this paper, referring to the grasp detection network framework proposed by [3], ResNet-101 is adopted as the backbone network of the grasp detection part in the second stage, as shown in Figure 4. The advantage of the ResNet network is to solve the problem that the network model

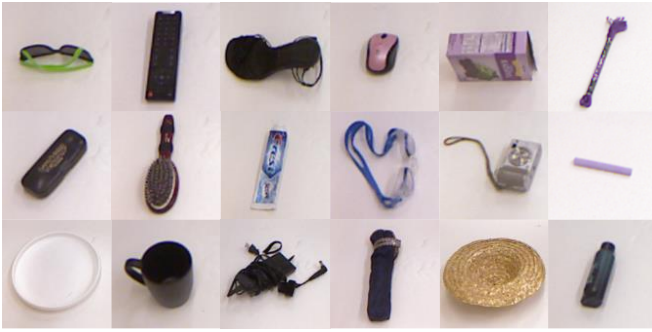

(a)

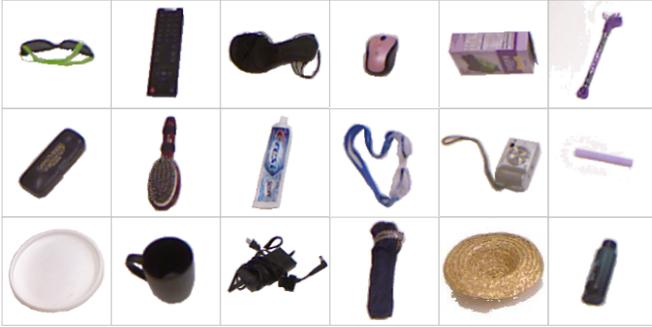

(b)

Figure 5 (a) Images from the Cornell Grasp Dataset and (b) images from the Cornell Grasp Dataset with background information removed.

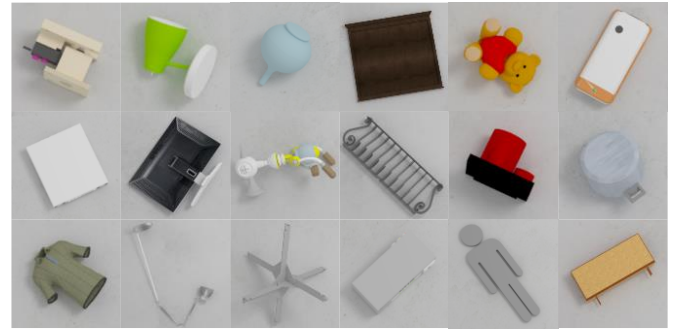

(a)

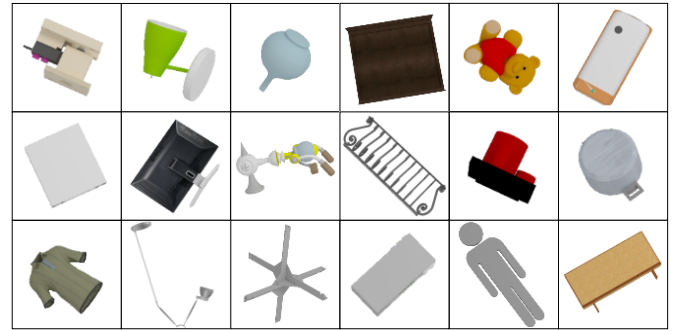

(b)

Figure 6 (a) Images from the Jacquard dataset and (b) images from the Jacquard dataset with background information removed.

degrades with depth employing residual mapping. This effective method enables the deep neural network to have optimal performance.

Different from the direct regression method proposed by [1][2] to predict a single 5-dimensional grasp representation $g = \{x, y, w, h, \theta\}$ in RGB or RG-D images. By quantifying angular data, our model divides angular velocity data into $R = 19$ categories at equal intervals and maps the calculation of angular data in the model into classification tasks. Simultaneously, to further improve the accuracy of classification, the classification of background elements is added to the above angle categories. In predicting model angle value, only when the probability of an angle category is higher than the probability of the background elements can be used as a credible grasping suggestion. Otherwise, the grasping suggestion is considered to be incorrect. Therefore, the total number of categories is $C = R + 1$. Our model predicts the grasp box data $\{x, y, w, h\}$ employing regression.

We use $\{C_1, C_2, C_3, C_4\}$ to represent the output characteristics of the residual network in various stages of the ResNet101 network. Among them, Grasp Proposal Network as Regional Proposal Network (RPN) [18] shares the output characteristics of the stage $C_3$ in ResNet101. Grasp the Proposal Network outputs a 1×1×512 feature map and fed it into two full connection layers. The outputs of the two full connection layers are 2×k and 4×k vectors (k is the number of anchor points), respectively, which respectively represent the possibility of grasping the position and the position representation of grasping the rectangle. The ROI layer extracts the suggested candidate box area features and outputs them to the subsequent network.

Grasp Proposal Network uses rectangular boxes with three scales and three aspect ratios to represent possible grasp positions for each anchor point in the feature map. Therefore, each input image generates a total of K×3×3 grasp location predictions. For the ground truth, we convert the grasp box with an angle into the form of $\{\theta, x_{\min}, y_{\min}, x_{\max}, y_{\max}\}$, where $\theta$ represents the angle of the grasp rectangle, and $x_{\min}, y_{\min}, x_{\max}, y_{\max}$ represents the horizontal grab rectangle's coordinates in the upper left corner and lower right corner of the image. This representation can be converted to a 5-dimensional grasping representation. We use $b_i$ to represent the horizontal grasping of the $i$-th 4-dimensional rectangle $(x_{\min}, y_{\min}, x_{\max}, y_{\max})$, and $p_i$ to represent the probability that the rectangle will be successfully grasped. For all grasp proposed $I$, we define the loss function of grasp proposes network as follows:

$$L_{gpn}(\{(p_i, b_i)_{i=1}^I\}) = \sum_i L_{gp\_cls}(p_i, p_i^*) + \lambda \sum_i p_i^* L_{gp\_reg}(b_i, b_i^*)$$

(2)

The loss function of the grasp proposal network is composed of two parts, in which $L_{gp\_cls}$ is the cross-entropy loss of grasp proposal classification and $L_{gp\_reg}$ is the $l_1$ regression loss with weight $\lambda$. At the same time, we define $p_i^*$ as the ground truth that can be grabbed. There are only 0 and 1 values. A probability of 0 means that the position can not be grasped, and a probability of 1 means that the position can be grasped. The variable $b_i^*$ represents the ground truth of the location grasped by the corresponding $p_i^*$.

The final stage of the model is to determine the configuration of the candidate grasping. At this stage, to effectively process the proposed area, we add the ROI pooling layer into our backbone network so that it can share the output feature map of the ResNet convolutional layer. The ROI pooling layer will concatenate all identified features of grasp proposal and shared features and then pass them into the two full connection layers for angle $\theta$ classification and anchor box parameter offset regression.

We use $\rho_\theta$ to represent the possibility that the angle will be $\theta$ after passing through the softmax layer and $\beta_\theta$ to describe the anchor corresponding to this Angle. Formula 3 accurately represents the grasp configuration prediction loss function.

$$L_{gcr}(\{(\rho_\theta, \beta_\theta)\}_{c=0}^C) = \sum_c L_{gcr\_cls}(\rho_\theta) + \lambda_2 \sum_{i=1} 1_{c \neq 0}(c) L_{gcr\_reg}(\beta_c, \beta_c^*) \quad (3)$$

Where $L_{gcr\_cls}$ is the cross-entropy loss of grasp Angle classification, $L_{gcr\_reg}$ is the $l_1$ regression loss with weight $\lambda_2$, which is used for the prediction of grasp anchor box, and $\beta_c^*$ is the ground truth of grasp anchor box position.

In the improved ResNet-101 model, the total loss adopted in the end-to-end training of grasp detection and grasp parameter prediction is $L_{total}$.

$$L_{total} = L_{gpn} + L_{gcr} \quad (4)$$

In general, our model extracts features through the ResNet-101 backbone network generates grasp proposals through the ROI layer. Finally, outputs grasp angle probability, grasp anchor frame parameters, and grasp credibility information through two full connection layers.

*C. Relationship with Previous Grasp Detection Algorithms*

At present, many grasp detection works [1][2] [3][4][15] have applied a deep neural network to their own models. The performance and accuracy of grasp detection are dramatically improved by successfully applied the network such as ResNet, Fast-RCNN, and VGG to grasp detection. These work by taking an RGB, RGD, or RGB-D image as input to the model and calculating potential grasp locations that may exist over the entire image. The ROI-GD grasp detector proposed by [5] takes the features of the ROI part of the object as input for grasp detection. This method significantly reduced the detection range, however, due to the object's irregularities, there is still a certain amount of irrelevant information in the ROI. Therefore, the previous detection methods cannot realize object-oriented grasp detection.

The method we propose is significantly different from the above methods. The grasp detection network we proposed uses the method of instance segmentation to segment the target object from the background and inputs the image containing only the pixel characteristics of the target object into the grasp detection network. The grasp detector takes the object's characteristics as input, which is more reasonable. Because what we are going to do is grasp detection for the target, not for the scene.

I. EXPERIMENTS AND EVALUATION

*A. Dataset*

We evaluated our model's performance in two different datasets: the Cornell Grasp Dataset and the Jacquard Dataset.

1) Cornell Grasp Dataset: The Cornell Grasp Dataset consists of 885 images with 244 different objects, and each object appears in the image in different positions and postures. Each image is marked with multiple ground truth grasp points, corresponding to the possible ways to grasp the target.

2) Jacquard Dataset: The Jacquard Dataset contains 54K images, 11k objects, and grasp annotations for 1 million locations. Compared with the Cornell Grasp Dataset, the Jacquard Dataset had more grasping annotations and was more reasonable.

During the experiment, we divided the dataset into a training set and a testing set according to the ratio of 4:1. Meanwhile, to train the segmented network of the model, we used the LabelMe labeling tool to label the training set of Cornell Grasp Dataset with mask label. The Cornell Grasp dataset was expanded 125 times to reduce the risk of overfitting, utilizing data enhancement such as translation, rotation, and brightness change.

*B. Pre-training*

To reduce the risk of overfitting during the learning process, our model started training based on pre-trained ResNet-101. As shown in Figure 4, we realized the grasp proposal's extraction by sharing the feature maps of the first three residual blocks in ResNet-101. These proposals are then fed into the ROI pooling layer and then fed into the fourth residual module of ResNet-101. The average pooling layer's output is fed into the two full connection layers for final classification and regression. Except for the backbone network ResNet-101, all new layers are trained from scratch.

As mentioned above, we transform grasp detection into two tasks: Angle classification and position regression. Therefore, we converted the Cornell Grasp Dataset's grasp annotation into the output format of the proposed network. We divided the continuous angle region from -90° to 90° into R regions on average and transformed the Angle ground truth into discrete categories.

*C. Metrics*

For the model proposed by us, in both the single-object and multi-object scenarios, the grasp detection phase is for one object. Therefore, to compute the accuracy of the grasp detection, we take the same metric "rectangular metric" widely used in all previous works. The grasp prediction configuration is correct if both:

1) the difference between the prediction grasp angle and ground truth is within 30°, and

2) the Jaccard index between prediction grasp $g_p$ and ground truth $g_t$ is greater than 0.25.

The calculation method of the Jaccard Index is:

$$J(g_p, g_t) = \frac{g_p \cap g_t}{g_p \cup g_t} \quad (5)$$

We evaluate our model with two different data split methods.

1) **Image-wise split**: splits datasets randomly. Dataset split is not based on categories, and each image is equivalent. Image-wise split tests the predictive ability of the model for the object in different locations.

TABLE I. ACCURACY OF DIFFERENT METHODS IN CORNELL GRASP DATASET

| Approach | Accuracy* | | Speed |
|---|---|---|---|
| | IW(%) | OW(%) | （FPS） |
| lenz et al.[13] | 73.9 | 75.6 | 0.07 |
| Redmon et al.[1] | 88.0 | 87.1 | 3.31 |
| Kumra et al.[2] | 89.2 | 89.0 | 16.03 |
| Guo et al.[15] | 93.2 | 89.1 | - |
| Asif et al.[19] | 90.6 | 90.2 | - |
| Fu-Jen et al. (RGB)[3] | 94.4 | 95.5 | 8.33 |
| Fu-Jen et al. (RGD)[3] | 96.0 | 96.1 | 8.33 |
| Zhang et al. [5] | 93.6 | 93.5 | 25.16 |
| **Ours: (RGB)** | **95.9** | **96.1** | **9.43** |
| **Ours: (RGD)** | **96.4** | **96.5** | **9.43** |

2) **Object-wise split**: splits object instances randomly. Categories in the test dataset do not appear in the training dataset. Object-wise split tests the generalization ability of the model for different objects.

## II. RESULTS

### A. Results for Our Proposed MASK-GD on The Cornell Grasp Datasets

We do not do any data augmentation for the data used to test the model. Moreover, the test images are input models one by one, rather than being processed in batches. The experimental results on Cornell Grasp datasets are shown in TABLE I. Our model achieves 96.4% and 96.5% accuracy in the image-wise split and object-wise split, respectively. Compared to the Baseline [3], our method improves accuracy by 1.5% and 0.6%, respectively, when inputting RGB images and 0.4% and 0.4%, respectively, when inputting RGD images. Our method eliminates the image's background elements that significantly affect the grasp detection performance and contrasts with the input image increase. On the other hand, replacing the value of blue channel pixel in RGB image with depth information also increases the contrast to some extent due to the difference in depth value between object and background. Therefore, in terms of precision improvement, the RGD format inputs are smaller than those in RGB format relative to baseline[3].

### B. Results for Our Proposed MASK-GD on The Jacquard Datasets

For the Jacquard dataset, the experimental results are shown in TABLE II. The accuracy of the MASK-GD method was 97.6%, which was 3% higher than that of ROI-GD[5]. Therefore, the performance of the MASK-GD method is higher than that of the previous state-of-the-art grasp detection algorithm[5].

It is not difficult to see that MASK-GD performance is different between the Cornell Grasp Dataset and the Jacquard Dataset. The accuracy of MASK-GD on the Jacquard dataset is better than that on the Cornell Grasp dataset. The Jacquard Dataset has dense grasp location annotations, mostly avoiding false predictions that can be grasped but do not meet the metrics. Therefore, our model has lower bias on the Jacquard Dataset than on the Cornell Grasp Dataset.

TABLE II. ACCURACY OF DIFFERENT METHODS IN JACQUARD DATASET

| Algorithm | Input | Accuracy (%) |
|---|---|---|
| Depierre et al. [20] | RGB-D | 74.2 |
| Zhou et al.[4] | RGB | 91.8 |
| | RGD | 92.8 |
| Zhang et al.[5] | RGB | 90.4 |
| | RGD | 93.6 |
| **Ours** | **RGB** | **97.1** |
| | **RGD** | **97.6** |

TABLE III. ACCURACY RATE AFTER CHANGING THE SCENE

| Algorithm | Accuracy on the Jacquard Datasets (%) | Testing accuracy on the Cornell Grasp Datasets (%) |
|---|---|---|
| Grasp(tradition) | 93.6 | 89.4 |
| **Seg + Grasp(ours)** | **97.1** | **96.8** |

### C. The Robustness of The Model to Background Information

In the current model's training and testing process, the model learns the whole scene's characteristics. Therefore, in the final grasp position prediction stage, the background features will impact the model's prediction results. As a result, the predicted performance of the model will be biased when the scene changes.

As shown in TABLE III, different models have different grasping performance changes after changing the grasping scene. For the traditional grasping model, the grasping detection model learns many correlation information between background and object in the training process. Therefore, when there is a big difference between the training scenario and the test scenario, the model's performance will deteriorate. Since the background elements are removed, the grasping detection model based on object segmentation proposed in this paper eliminates the interference of background information on grasping detection. Therefore, the performance of the grasping model is not affected when the model usage scenarios change.

## III. CONCLUSION

We propose a new method Mask-GD for robot grasping detection. This method eliminates background information interference to the grasping detection by object segmentation and realizes target detection instead of the scene. We verified our proposed methods on the Cornell dataset and Jacquard dataset, respectively, and compared them with the state-of-the-art system using a standard performance metric and methodology to demonstrate our methods' effectiveness. The experimental results show that our proposed method outperforms the state-of-the-art system.